\documentclass[pdflatex,sn-mathphys-ay]{sn-jnl}

\usepackage{graphicx}%
\usepackage{multirow}%
\usepackage{amsmath,amssymb,amsfonts}%
\usepackage{amsthm}%
\usepackage{mathrsfs}%
\usepackage[title]{appendix}%
\usepackage{xcolor}%
\usepackage{textcomp}%
\usepackage{manyfoot}%
\usepackage{booktabs}%
\usepackage{algorithm}%
\usepackage{algorithmicx}%
\usepackage{algpseudocode}%
\usepackage{listings}%

\theoremstyle{thmstyleone}%
\theoremstyle{thmstyletwo}%

\theoremstyle{thmstylethree}%

\begin{document}

\title[Article Title]{Increasing the Accessibility of Causal Domain Knowledge via Causal Information Extraction Methods: A Case Study in the Semiconductor Manufacturing Industry}

\author*[1,2]{\fnm{Houssam} \sur{Razouk}  } \email{houssam.razouk@infineon.com}
\equalcont{These authors contributed equally to this work.}
\author[1,3]{\fnm{Leonie} \sur{Benischke} }\email{benischke.leonie@gmail.com}
\equalcont{These authors contributed equally to this work.}
\author[2]{\fnm{Daniel} \sur{Gärber}}\email{danielgaerber8@gmail.com}
\author[2,4]{\fnm{Roman} \sur{Kern}}\email{rkern@tugraz.at}

\affil*[1]{\orgdiv{Quality Management}, \orgname{Infineon Technologies Austria AG}, \orgaddress{ \city{Villach}, \postcode{9500}, \state{Carinthia}, \country{Austria}}}

\affil[2]{\orgdiv{Institute of Interactive Systems and Data Science}, \orgname{Graz University of Technology}, \orgaddress{\city{Graz}, \postcode{8010}, \state{Styria}, \country{Austria}}}
\affil[3]{\orgdiv{Centre de Recherches Interdisciplinaires et Transculturelles}, \orgname{Université Bourgogne Franche-Comté}, \orgaddress{\city{Besançon}, \postcode{25000}, \state{Franche-Comté}, \country{France}}}
\affil[4]{\orgname{Know-Center GmbH}, \orgaddress{\city{Graz}, \postcode{8010}, \state{Styria}, \country{Austria}}}


\abstract{
The extraction of causal information from textual data is crucial in the industry for identifying and mitigating potential failures, enhancing process efficiency, prompting quality improvements, and addressing various operational challenges.
This paper presents a study on the development of automated methods for causal information extraction from actual industrial documents in the semiconductor manufacturing industry. 
The study proposes two types of causal information extraction methods, single-stage sequence tagging (SST) and multi-stage sequence tagging (MST), and evaluates their performance using existing documents from a semiconductor manufacturing company, including presentation slides and FMEA (Failure Mode and Effects Analysis) documents. 
The study also investigates the effect of representation learning on downstream tasks. 
The presented case study showcases that the proposed MST methods for extracting causal information from industrial documents are suitable for practical applications, especially for semi structured documents such as FMEAs, with a 93\% F1 score. Additionally, MST achieves a 73\% F1 score on texts extracted from presentation slides.
Finally, the study highlights the importance of choosing a language model that is more aligned with the domain and in-domain fine-tuning.

}


\maketitle

\section{Introduction}\label{Introduction}

Causal domain knowledge plays a vital role in various downstream tasks, such as risk assessment~\citep{hu2013software}, root cause analysis~\citep{saha2022mining} and data mining~\citep{anand1995role}.
The smallest unit of the causal domain knowledge is known as a causal relationship.
The causal relationship is a connection between two or more events or variables where one event or variable is the cause of the other.
Causal domain knowledge is commonly documented either in unstructured or semi structured forms.
Tabular formatted documents, typically utilized in the Failure Mode Effect Analysis (FMEA), serve as a prime example of semi structured documents.
Furthermore, much causal domain knowledge is also found in presentation slides, 
which are documents that do not follow a predefined structure, and are typically used in industry for sharing information between the different teams and with customers.

In FMEA documents, the manually created textual content presents several challenges such as non-standardized descriptions of the failure modes, effects, and root causes, and in many cases merged cells. 
In the case of merged cells, the description of multiple events within a single FMEA table cell~\citep{razouk2022improving}, which can involve enchained relations.
In an enchained causal relation, a cause or an effect in a causal relation is the cause or the effect of another causal relation. 
This creates a complex network of interconnected events where each cause and effect is both influenced by and influences other events in the chain.

Furthermore, the increasing use of digital presentation slides as a medium for presenting and sharing information has made them a valuable source of knowledge~\citep{hayama2008structure}. 
Presentation slides stemming from processes like failure analysis requests encapsulate documented causal relations. 
These relations shed light on failures detected in products, providing a comprehensive understanding of the factors contributing to such issues.
However, the unstructured nature of presentation slides, which combine visual and textual elements and use spatial positioning to document information, can make automated information access difficult.
While these features are useful for presenting information to a live audience, they can make it difficult for automated systems to extract and understand the information contained in the slides.

Presentation slides and FMEA documents are a rich source of causal domain knowledge. 
However, the sheer volume of documents in the industry can make it difficult to manually process and extract the information contained within them.
This can lead to increased product development cycle time, which can be detrimental to overall productivity. 
Therefore, automatically extracting causal domain knowledge from unstructured documents like presentation slides, as well as from semi structured documents like FMEA documents, can be highly beneficial in increasing the availability and accessibility of this knowledge.

Scholars have been addressing the topic of causal information extraction from text devising natural language processing (NLP) methods. 
Two common NLP approaches are lexical pattern-based methods~\citep{girju2003automatic} and statistical machine learning-based methods~\citep{yang2022survey}.
More recently, pre-trained transformers based language models are devised for extracting meaningful representation of the text in sequence tagging based approach for extracting causal information~\citep{gaerber2022causal,Text-to-Causal-Knowledge-Grap,saha2022spock}.
Also, research indicated that fine-tuning on domain specific data can enhance the vector text representation by better capturing the nuances and complexities of the domain-specific language, which can improve the overall performance on downstream tasks \citep{lee2020biobert}.

Even though causal information extraction has become a well studied topic in recent years,
streamlined causal information extraction methods tend to struggle with the detection of detailed causal relations including nested and enchained relations~\citep{saha2022spock,Text-to-Causal-Knowledge-Grap}.
Furthermore, testing of these methods on different types of industrial documents is limited.
To address this gap, the following research questions are addressed in this paper:

\begin{itemize}

    \item How effective are existing causal information extraction methods on different types of industrial documents?
    \item What is the effect of text representation learning on the overall performance of causal information extraction in industrial settings?
    
\end{itemize}

In summary, the aim of this research is to develop a method for detailed causal information extraction on industrial data, with which the availability and consistency of causal domain knowledge can be improved in the industrial world, facilitating more reliable data analysis and more informed decision-making.
Our work has the potential to be applicable to a variety of domains, including teaching, where much causal domain knowledge is documented in unstructured or semi structured documents.
By addressing these research questions, the contribution of this paper can be summarized as follows:

\begin{itemize}
    \item Extending causal information extraction methods to industrial documents,  increasing the availability of causal domain knowledge for downstream tasks in the industry.
    \item Providing guidance for practitioners working in different industries with similar types of documents.
    \item Addressing data consistency issues commonly found in semi structured documents, like the merged cells in FMEA (Failure Mode and Effects Analysis) documents.
    \item Contributing to the body of research that highlights the effect of representation learning on downstream tasks.
\end{itemize}

\section{Related Work}\label{Related Work}

The extraction of causal information from documents is a widely discussed topic across various domains such as the medical, financial, and industrial sectors. The importance of extracting causal information lies in its various applications, which are summarized in Section~\ref{CIE applications}.
Scholars have been actively developing different methods to extract causal information from documents due to the high demand for this task. These methods can be broadly categorized into two groups: knowledge-based methods and data-driven methods. Section~\ref{CIE methods} provides a summary of these methods and their respective approaches. 
Overall, the extraction of causal information is a critical task that has numerous applications across various domains, and researchers continue to develop innovative approaches for its efficient extraction.

\subsection{Causal information extraction applications}
\label{CIE applications}

Causal information extraction plays a crucial role in diverse sectors. 
The interest in this topic is especially high in the health~\citep{richie2022inter}, finance~\citep{saha2022spock} and industrial~\citep{RazoukIEEE} sector.
Causal information extraction in the health sector is essential for advancing medical knowledge~\citep{reklos2022medicause}, improving patient care~\citep{seol2014problem}, supporting research endeavors~\citep{negi2019novel}, and enhancing overall healthcare outcomes. 
It plays a critical role in various aspects of healthcare, from disease understanding and diagnosis to treatment personalization, drug discovery, and public health management.
Causal information extraction in finance can enhance risk management \citep{aerts2014management}, market analysis \citep{nam2019financial}, fraud detection~\citep{kong4516310cftnet}, portfolio optimization~\citep{tang2019global, ravivanpong2022towards}, regulatory compliance \citep{ravivanpong2022towards}, policy development, and scenario analysis. Causal information empowers financial professionals with valuable insights, assisting them in navigating the complexities of the financial landscape. 
Extensive research in this field has been conducted as part of the shared task FinCausal 2022~\citep{mariko2022financial}.

In industrial settings, ontologies are commonly employed to create a knowledge base and facilitate the sharing of knowledge in a structured manner.
As shown in ~\citep{safont2021using}, they can also be used to structure causal domain knowledge.
However, ontological approaches often encounter challenges related to high maintenance requirements and limited scalability. 
To address these issues, considerable effort has been invested in automatically enriching tabular formats with additional information derived from domain knowledge. 
Razouk and Kern \citep{razouk2022improving} present approaches that use statistical pattern-based methods combined with lexical patterns to detect cases of merged cells, aiming to enhance the consistency of these documents.
Furthermore, in the realm of risk assessment documents, such as FMEA documents, causal domain knowledge has been extracted and structured into a knowledge graph~\citep{RazoukIEEE}.
This knowledge graph has been ingeniously employed to develop a method for knowledge discovery, drawing from common sense knowledge completion techniques

\subsection{Causal information extraction methods}
\label{CIE methods}

The extraction of causal information from documents is a topic of great interest since not only numerical data but also texts can give valuable insight into causal relations~\citep{yuan2023tc}. 
While causal discovery aims to uncover the causal model or, at the very least, a Markov equivalent that mimics the data generation process for a given data set~\citep{glymour2019review},  causal information extraction from texts focuses on identifying causal entities and how they are connected to each other~\citep{yang2022survey}.
The current existing methods for causal information extraction are broadly categorized by~\citep{yang2022survey} into three main categories:
\begin{enumerate}
    \item  Pattern-based approaches
    \item  Statistical machine learning-based approaches 
    \item  Deep learning approaches 
\end{enumerate}

Pattern-based approaches usually leverage linguistic patterns to identify causal language in limited contexts. 
For instance, ~\citep{girju2003automatic} use lexical and syntactic patterns to detect causal relations in medical and business texts. 
Pattern-based approaches can be effective for specific domains but they lack generalizability.
Statistical machine learning-based approaches typically involve the use of third-party NLP tools which require fewer patterns than pattern-based approaches~\citep{wu2012detecting}.
Deep learning approaches leverage deep neural networks to extract causal information from text~\citep{akkasi2021causal} and to acquire useful vectorized representations of text~\citep{gaerber2022causal,saha2022spock,Text-to-Causal-Knowledge-Grap}. 
This involves training neural networks to identify causal relationships in text using techniques such as supervised learning.
As such, many scholars have been working on creating annotation guidelines~\citep{dunietz-etal-2015-annotating,dunietz-etal-2017-corpus,DBLP:journals/corr/abs-2012-02498} and annotating data sets for training and testing the developed methods~\citep{mariko2022financial,narduzzi2022S2ORC-SemiCause}.

One of the most common machine learning and deep learning approaches for extracting causal information is the use of sequence tagging.
As causal information extraction is not only about identifying causes and effects, but about detecting them as pairs, to obtain useful causal information, the relation between the entities needs to be taken into consideration~\citep{yang2022survey}. 
Pure sequence tagging methods, such as the method presented in~\citep{saha2022spock,khetan2020causal}, are facing difficulties in integrating relational information into the input data while performing Named Entity Recognition (NER).
Since the relation between the entities is crucial in order to obtain useful causal knowledge, other approaches have been proposed which involve fine-tuning pre-trained language models for text span classification and sequence labeling tasks. 
For example, the authors in~\citep{saha2022spock} specifically label the text spans corresponding to cause and effect in a given text. They then proceed to classify whether these identified cause and effect spans were linked together through a causal relation. 
Similarly,~\citep{khetan2020causal} employ an event-aware language model in order to predict causal relations by taking into account event information, sentence context and masked event context. 
Another major difficulty in the extraction of causality using NER is the recognition of overlapping and nested entities. 
~\citep{lee2022mnlp} tackle overlapping entities by employing Text-to-Text Transformer (T5).
Nevertheless, profound nested relations in which whole causal relations are nested within an entity are, however, not captured with their method.
Also, Gärber~\citep{gaerber2022causal} has proposed a multi stage sequence tagging (MST) approach for extracting causal information from historic texts. 
MST method extracts causal cues in the first stage and then uses this information to extract complete causal relations in subsequent stages.

To the best of our knowledge, the testing and adaptability of methods for causal information extraction from industrial documents remain relatively limited. 
Furthermore, there is a lack of comprehensive exploration regarding the challenges faced by such methods concerning different document formats, such as presentation slides and tabular formats, as well as regarding domain-specific language in comparison to general language.
Research in these aspects could provide valuable insights and guidance for the development of information extraction methods from industrial documents.

\section{Method}\label{Method}

\begin{figure*}
  \centering
  \includegraphics[width=\textwidth]{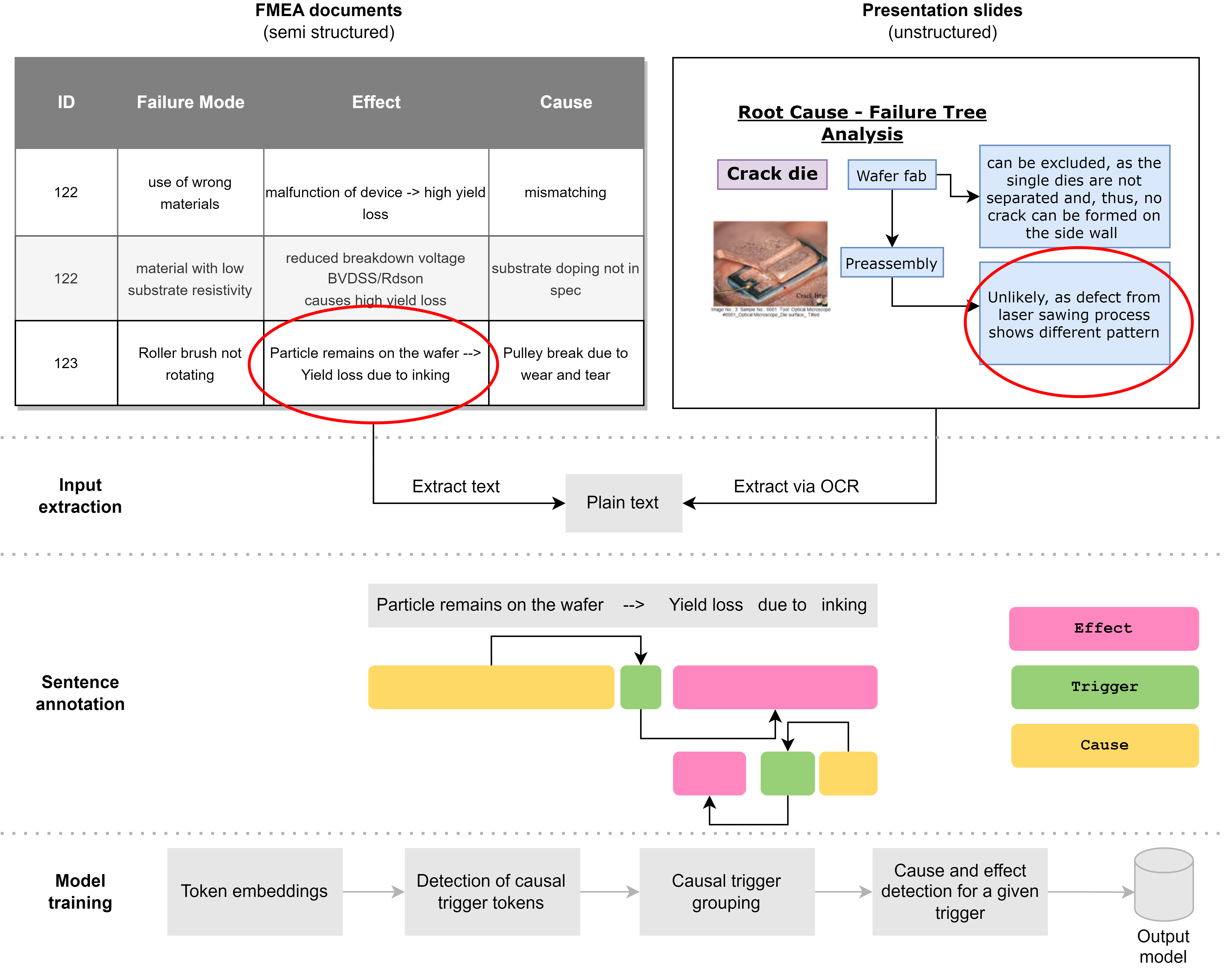}
  \caption{\textbf{Proposed method for causal information extraction from various industrial documents.}
  For the extraction of causal information, texts are gathered from tabular FMEA documents and industrial presentation slides.
  The causal entities and relations in these texts are then annotated following specific annotation guidelines, as described in Section~\ref{subchap:Annotation}.
  The depicted example illustrates a text extracted from and FMEA cell which contain two causal relations.
  For higher generalizability, meaningful text representation is generated using various language models. 
  This representation is then used to train different sequence tagging models for causal information extraction. 
  Namely multi-stage sequence tagging approach which uses a cascade of models for different tasks is depicted.}
  \label{fig:pipeline}
\end{figure*}

Causal domain knowledge is typically documented either in unstructured or semi structured documents.
These documents are by design aimed to be produced and consumed by human domain experts.
Thus, this documentation style impedes a straight forward automated information access. 
The objective of this study is to develop a method that can increase the usability of the causal domain knowledge contained in a set of industrial documents by transforming the information into a more structured format.
To achieve this objective, we propose a causal information extraction method from various industrial documents, depicted in Figure~\ref{fig:pipeline}, that involves the following steps:

\begin{enumerate}

    \item Text extraction from different documents formats: this step involves extracting a textual representation of the information contained in the different documents.
    This textual representation can be analyzed and used to develop causal information extraction methods.
    This step is elaborated in Section~\ref{Text extraction from different documents formats}.

    \item Structured Data Representation: This step involves representing the information contained in texts extracted from different industrial documents in a structured format (i.e. a set of named entities and relations between these entities) that can be easily used for downstream tasks such as risk assessment, data analysis, etc. This step is outlined in Section~\ref{subchap:Annotation}.

    \item Automated causal information extraction from text: This step involves extracting meaningful vectorized text representation.
    Leveraging this representation to develop causal information extraction methods that transfer the information contained in the text to a set of named entities and relations between these entities. Namely, we propose two approaches: the first one based on a single stage sequence tagging (SST), the second one is based on multi stage sequence tagging (MST). This step is elaborated in  Section~\ref{subchap:Automated CIE}.
    
\end{enumerate}

\subsection{Text extraction from different documents formats}
\label{Text extraction from different documents formats}

Our method focuses on two types of industrial documents: semi structured documents represented by FMEA documents, and unstructured documents represented by presentation slides. 
For semi structured documents, our method parses the documents as tabular formats and extracts the text contained in the failure mode, effect, and cause columns. 
Each cell of these columns represents a separate entity and is treated as a standalone text.
This approach has also been proposed by Razouk et al. in their work on FMEA documents~\citep{razouk2022improving,RazoukIEEE}. 
By extracting causal information from FMEA text, merged cells which contain causal relations can be split into multiple cells and merged into the existing knowledge graph.
Thus,a more connected knowledge graph can be achieved.

To extract the data from the presentation slides files, the slides are converted to images and then read by an OCR architecture composed of the text detection model and the text recognition model. 
This method also offers to recognize each stand alone text based on its position in regard to other texts in the slide.
The selection of this method is attributed to the fact that, to date, methods for text extraction directly from presentation slides files are still lacking in term of reliability.
Even though there are libraries for this purpose, such as python pptx, they do not offer support for extracting content from graphic frames, which includes SmartArt.
Furthermore, these libraries extract the content of the slides in the order in which the boxes were created and not in the order in which they are intended to be perceived.

\subsection{Structured data representation of causal information}
\label{subchap:Annotation}

Structured data representation is a crucial part of information extraction. 
In particular, representing causal relationships in a structured manner have received significant attention in recent years due to their importance in understanding the connections between different events, phenomena, and concepts.
There are several ways to represent structured data in the context of causal relationships. 
One common approach is to use knowledge graphs to represent the relationships between different entities.
In these graphs, each entity is represented as a node, and the causal relationships between entities are represented as edges. 
For example, a knowledge graph could represent the causal relationships between different failure modes and their causes and effects in the FMEA documents~\citep{RazoukIEEE}.

However, information extraction for causal relations is challenging due to several reasons.
Causal relations are often part of our cognitive reasoning and can be articulated implicitly in text, making it difficult for automated systems to identify them accurately.
Additionally, cognitive reasoning can differ from person to person, which can create disagreement in the interpretation of the same text. 
To mitigate these challenges, several scholars have proposed different annotation guidelines to guide annotators in the annotation process~\citep{dunietz-etal-2015-annotating,dunietz-etal-2017-corpus,DBLP:journals/corr/abs-2012-02498,khetan2021mimicause}.
Specifically, many scholars have recommended only extracting explicit causal relations \citep{dunietz-etal-2015-annotating}. 
In our structured data representation, we have adapted these recommendations by specifying three entity types: cause, effect, and trigger. 
The cause and effect entities represent events or variables articulated in the text, while the trigger is the explicit causal clue that expresses the presence of the causal relation between the cause and the effect.
To fully represent a causal relation, three types of entities - cause, trigger, and effect - and two relations between them are used. The first relation connects the cause and the trigger, while the second relation connects the trigger and the effect.
This approach helps to accurately capture explicit causal relations in texts and can be useful in various applications such as information retrieval and knowledge representation.

Annotation guidelines are crucial for creating consistent training and testing data, particularly for data-driven methods. 
They facilitate the consistency required for effective training and evaluation of such methods. 
However, these guidelines may have limited applicability to domain-specific texts, such as those from the semiconductor industry, due to their unique characteristics.
To address these challenges, we have devised annotation guidelines that combine and distill the diverse paradigms posited by prior researchers.
Our guidelines provide examples from the domain-specific data set to support experts in gaining a better grasp of the task. 
The developed annotation guidelines are detailed as follows:

\begin{enumerate}

    \item  Causal relations are only annotated on text level, which corresponds to a single FMEA cell or a text box recognized by the OCR in the case of a presentation slides.
            Relations between entities that belong to different texts are disregarded.
    \item  Two entities - either two effects or two causes - are annotated as a single entity, if they are 
            linked to the same cause or effect~\citep{DBLP:journals/corr/abs-2012-02498}.
            Example: \textit{\textbf{Die chipping/crack}\textsubscript{Effect} \textbf{due to}\textsubscript{Trigger}\textbf{dicing process condition/parameters and the wafer condition in kerf area}\textsubscript{Cause}}.
    \item Causal relations can be chained~\citep{DBLP:journals/corr/abs-2012-02498}.
            The effect of a cause can be the cause of another effect.
            Example: \textit{\textbf{Due to}\textsubscript{Trigger} \textbf{a wrong implantation dose}\textsubscript{Cause},\textbf{the compensation was destroyed}\textsubscript{Effect} \textsubscript{Cause}, and \textbf{therefore}\textsubscript{Trigger}, \textbf{the lot was disregarded.}\textsubscript{Effect}}
            In the example, the entity “the compensation was destroyed” is the effect of “a wrong 
            implementation dose” and the cause of “the lot was disregarded.
    \item Nested relations are allowed.
    There can be causal relations inside an entity (cause or effect).
    Example: \textit{ \textbf{Foreign material or residue does not cause failure at wafer test}\textsubscript{Effect} \textbf{due to}\textsubscript{Trigger} \textbf{thin isolation, 
    inhibiting leakage current.}\textsubscript{Cause}.}
    In the example, “thin isolation, inhibiting leakage current” is the cause of the first part of 
    the sentence but within this cause, there is another causal relation, since “Thin isolation” 
    is the inhibitory cause of “leakage current”. In this case, we annotate the cause as well as 
    the entities and the relations within the cause.
    \item   Entities can be interrupted by other entities.
    Interrupted entities are annotated as one entity, excluding the part that belongs to another entity. 
    Example: \textit{\textbf{Due to a wrong implantation dose, the compensation was destroyed}\textsubscript{Cause}, and \textbf{the lot was}\textsubscript{Effect} \textbf{thus}\textsubscript{Trigger} 
    \textbf{disregarded}\textsubscript{Effect} .}
    \item   Entities are only annotated, if there is a complete causal relation with a cause, an effect 
    and a trigger.
    \item Causal relations without an explicit trigger are disregarded~\citep{dunietz-etal-2017-corpus}. 
    \item Lexical causatives are disregarded~\citep{dunietz-etal-2017-corpus,dunietz-etal-2015-annotating}.
    Example: \textit{Electrical and mechanical stress at application environment is cracking the isolation layer 
    between defect and conductive line.}
    Sentences with transitive verbs like “to crack” are not considered causal, even though one 
    could argue that, in the example, the action of cracking is the cause for the crack. 
    Such relations are disregarded, since the entities and the trigger cannot be clearly separated.
    \item    Vague causal relations are disregarded~\citep{khetan2021mimicause,dunietz-etal-2017-corpus}.
    Example: Sentences like “Y is linked to X” are not annotated. 
    \item Hypothetical and assumed causal relations are annotated.
     Example: \textit{\textbf{Scratches at Wafer BS}\textsubscript{Effect}, \textbf{most probably due to}\textsubscript{Trigger} \textbf{particles}\textsubscript{Cause}.} 
    \item Future causal relations are considered.
    Example: Sentences like “X will lead to Y” are annotated.
    \item Relative pronouns are annotated as part of the cause or effect.
    Example: \textit{There is a QMP regarding \textbf{edge damage which}\textsubscript{Cause} \textbf{could cause}\textsubscript{Trigger} \textbf{the flying dies}\textsubscript{Effect}.}

    \end{enumerate}

\subsection{Automated causal information extraction from text}
\label{subchap:Automated CIE}

The proposed approach utilizes BERT-based language models to extract meaningful vectorized text representations, specifically selected for their relevance to the domain. 
To potentially further optimize these models, in-domain fine-tuning using masked language modeling objective is conducted on a data set from the same domain. 
During the fine-tuning, we compare two masking objectives. 
The first objective involves uniform masking (UM), where the masked tokens are selected randomly across the text.
The second objective leverages point-wise mutual information masking (PMI)~\citep{levine2020pmi}, where spans of tokens related to each other based on their co-occurrence are masked. 
This approach could potentially help to improve the relevance of the extracted vectorized text representations for the specific domain. 
Hence, by selecting and fine-tuning BERT-based models, we could effectively capture the nuances and complexities of the domain-specific language and improve the overall performance 

\begin{figure*}
  \centering
  \includegraphics[width=\textwidth]{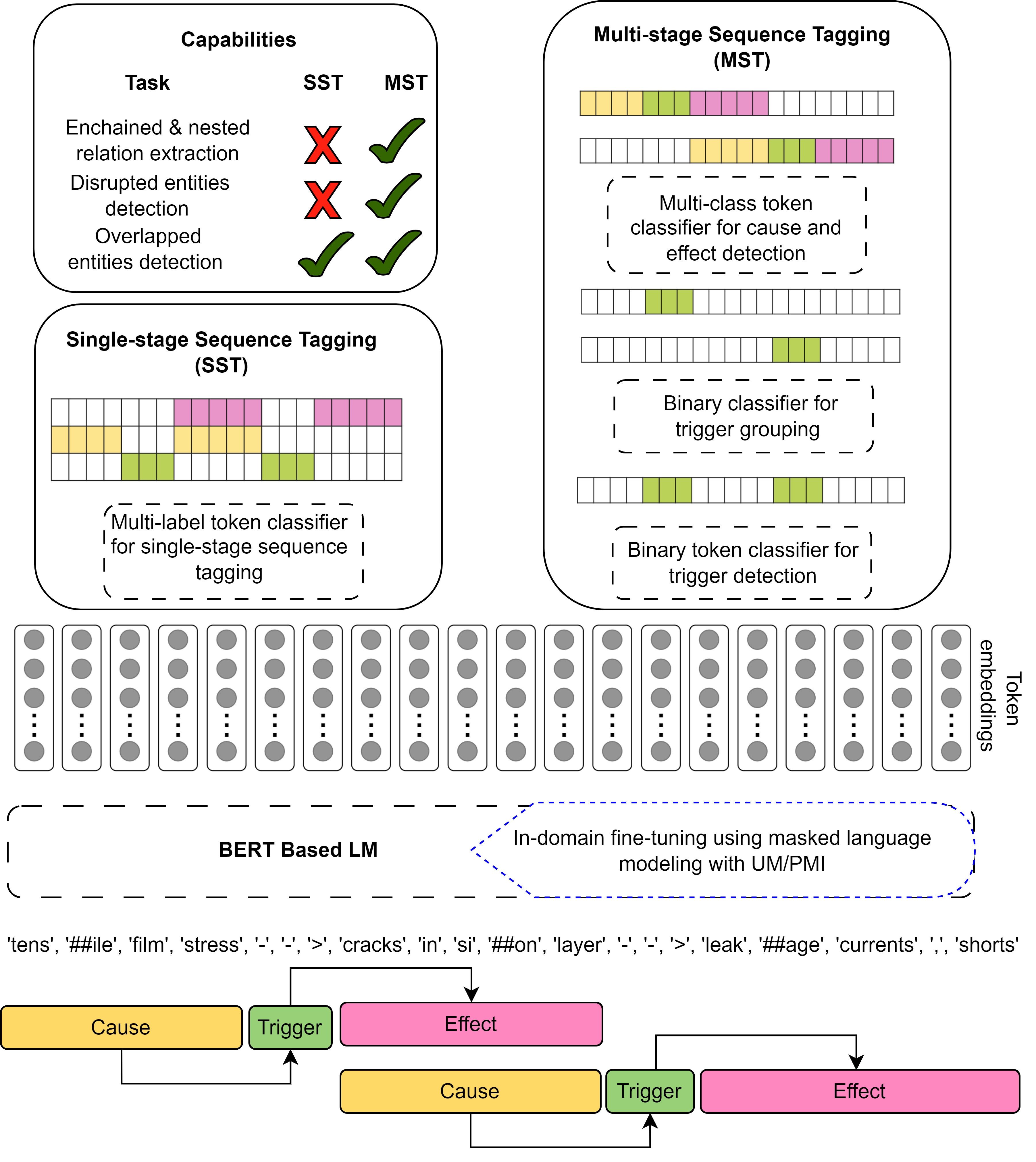}
  \caption{\textbf{Automated Causal Information Extraction from Text.} The proposed approach for causal information extraction from text utilizes a BERT-based language model that is selected based on the relevance of its initial training data set to the domain of interest. The model is initially fine-tuned for the domain using a masked language modeling objective, with two masking strategies (UM and PMI) being compared.
  A portion of the annotated data set is used to train the model, and two causal information extraction methods are compared: single-stage sequence tagging (SST) and multi-stage sequence tagging (MST).
  SST, based on multi-label token classification, is capable of detecting overlapped entities. 
  MST cascades multiple models, including a binary classifier for trigger detection, a binary classifier for trigger grouping, an attention network for trigger combined embedding, and a multi-label classifier for argument detection.
  In addition to detecting overlapped entities, MST is capable of extracting enchained relations and detecting disrupted entities.}
  \label{fig:CIE}
\end{figure*}

To transfer the information contained in text, two causal information extraction methods are selected.
The first approach is based on single-stage sequence tagging (SST) using BERT token embeddings, originally designed for named entity recognition.
Specifically, the proposed SST approach extracts the token embeddings from the text, and passes them to a multi-label classifier, which allows for detecting overlapping labels, meaning a token can be recognized as a cause and an effect at the same time. 
The losses of the classifier are used to update both the classifier weights and the language model weights. 
This approach is depicted in left side of Figure~\ref{fig:CIE}.

The second approach is based on a multi-stage sequence tagging (MST) method, firstly presented in \citep{gaerber2022causal}.
This approach consists of multiple cascaded models, that utilize the text embeddings given by a BERT model.
The first model is a binary token classifier that detects tokens with a trigger label. 
The output of this model is passed to a second model that identifies tokens belonging to the same trigger entity.
This is achieved using a trigger grouping model that performs binary classification to determine whether two token embeddings with predicted trigger labels belong to the same trigger entity. 

This model allows the method to detect disrupted entities which could be the case for the trigger.
Disrupted entities refer to entities that have been interrupted by the occurrence of another entity within its text span. 
In the sentence, "\textit{The root cause for back side defect are humidity residues}", the trigger is made up of the tokens "\textit{The root cause for}" and "\textit{are}". 
However, this trigger is disrupted by the effect, which is represented by the tokens "\textit{back side defect}".

In the next step, the embeddings of the tokens belonging to the same trigger entity are aggregated together using an attention network.
The aggregated combined trigger embeddings are devised by a third model to detect the arguments (i.e., tokens belonging to the cause and effect of that trigger). 
Specifically, the model concatenates the embeddings of the tokens and the combined trigger embeddings and devices them as an input to a multi-class calcification to detect the cause and effect for a given trigger. 
As such a token could be predicted to be a cause for a certain trigger and the effect of another trigger. 
As a welcomed result, the MST supports extracting enchained relations.

The model is trained in an end-to-end manner, and the weights are updated using a combined loss function.
This includes the weights of the language model, as well as the binary token classifiers for trigger detection and grouping, the attention network for trigger embeddings aggregation, and the multi-class classifier for detecting arguments. 
For each training example, the model processes the input data through its various components and produces an output. 
The output is then compared to the desired output, and a combined loss value is computed.
This approach is depicted on the right side of Figure~\ref{fig:CIE}.

\subsection{Evaluation Methodology}
\label{subchap:Evaluation}

To evaluate the performance of our proposed pipeline for causal information extraction in the semiconductor industry, a set of evaluation techniques is devised.
These techniques are detailed as follows.
To assess the clarity and effectiveness of the structured data representation of causal information, the consistency of annotations among multiple annotators is evaluated. 
This metric can help predict potential limitations in the clarity of the annotations guidelines that could propagate to the automated causal information extraction methods. 
To estimate the inter-annotator agreement (IAA), the token-based Cohen’s $\kappa$ and $F1$ score are calculated. 
While the performance of the method is not bound by the IAA~\citep{richie2022inter}, the IAA can set or limit expectations and indicate which aspects might be particularly challenging for the model to learn.
In addition, we executed a performance evaluation of each component of the MST pipeline, including trigger detection, trigger grouping, and cause and effect detection.
For each component, the $F1$ score is calculated, and the results between the two approaches (i.e. SST and MST) are compared.

\section{Experiments and Results}\label{Experiment and Results}

In pursuit of our research objectives, we collected a data set containing two types of industrial documents from an actual semiconductor manufacturing company.
The data set comprises a set of presentation slides and a set of Failure Mode and Effects Analysis (FMEA) documents. 
The text from the FMEA documents was extracted at the tabular cell level, similar to the work of Razouk et al.\citep{razouk2022improving} and\citep{RazoukIEEE}. 
Mainly cell texts belonging the failure mode effect and root cause columns are extracted. 
To extract the textual elements of the presentation slides in their intended order, an optical character recognition (OCR) based method is utilized.
Although OCR based method produces occasional errors, at this stage it represents the best option for purposes of this research.
Specifically, the \detokenize{db_resnet50} text recognition model and the \detokenize{crnn_mobilenet_v3_large} model from DocTR~\citep{doctr2021} are utilized.

The collection of texts from various presentation slides and FMEA documents underwent annotation using the Brat annotation tool~\citep{brat}.
A team of two proficient Natural Language Processing (NLP) experts, with a thorough understanding of the provided annotation guidelines, carried out the annotation process separately. 
The annotated data set comprises 495 texts from FMEA documents and 440 texts from presentation slides.
Cohen's $\kappa$ and $F1$ score were used to estimate the inter-annotator agreement (IAA), which was calculated on the token level of the text using multiple language model tokenizers. 
The selected language models were BERT~\citep{devlin2018bert} and MatBERT~\citep{walker2021impact}. 
These models are chosen because BERT is a solid baseline architecture that is used for many downstream general domain tasks, and MatBERT was specifically trained to handle materials science terminologies, which is relevant to the semiconductor manufacturing industry. 
Here, we make use of the data sets and the in-domain fine-tuning pip line, which were established by Tosone~\citep{Tosone2022Automatic} using semiconductor manufacturing related data set.
In addition to the annotated data set, an additional set of 481 texts extracted from presentation slides underwent annotation by only one NLP expert.

To gain further insights, the inter-annotator agreement (IAA) is estimated for each label type (Cause, Effect, and Trigger) and for each source of text represented in the data set (i.e., presentation slides and FMEA documents) separately. 
Table~\ref{tab:Inter_Annotater_Agreemnet} summarizes the results of the IAA for different data sets and tokenization of various language models.
Based on Table~\ref{tab:Inter_Annotater_Agreemnet}, the overall inter-annotator agreement (IAA) for all three label types is 85\% Cohen's $\kappa$.
However, substantial differences in the IAA can be observed when comparing different labels. 
While there is high agreement in the annotation of triggers with an IAA of 94\% Cohen's $\kappa$, the annotation of causes and especially the annotation of effects proves to be more challenging, with an IAA of 86\% and 76\% Cohen’s $\kappa$, respectively.
The IAA for texts from FMEA documents is generally higher than for texts from presentation slides.

\begin{table}[h]
\caption{
\textbf{Inter-annotator agreement results for different data sets.} 
The inter-annotator agreement (IAA) is higher for the data extracted from semi structured documents, namely FMEAs, compared to the data extracted from presentation slides.
Additionally, the IAA is higher for annotations of the 'Trigger' type compared to annotations of the 'Cause' and 'Effect' types. 
These results suggest that the extraction of causal information from FMEA documents is less ambiguous and more consistent compared to the extraction from presentation slides, and that the 'Trigger' type annotations are easier to identify and agree upon by annotators.}
\label{tab:Inter_Annotater_Agreemnet}
\begin{tabular}{ l c c c c c c c c}
\toprule
 \multirow{3}{*}{Annotation Type}& \multicolumn{4}{c}{Data set} &\multicolumn{2}{c}{Micro Avg} &\multicolumn{2}{c}{Macro Avg} \\
   &\multicolumn{2}{c}{FMEA}  &\multicolumn{2}{c}{Slides} &\multicolumn{2}{c}{FMEA\&Slides} &\multicolumn{2}{c}{FMEA\&Slides}\\
   &  $\kappa$\%  & $F1$\%  &  $\kappa$\%   & $F1$\%   & $\kappa$\%   & $F1$\%   & $\kappa$\%   & $F1$\%  \\
\midrule

 Trigger    & 94& 94 & 93& 93 & 94& 94   & 94& 94\\
 Cause      & 87& 90 & 84& 85 & 87& 88   & 86& 88\\
 Effect     & 86& 88 & 67& 68 & 81& 83   & 76& 78\\
 Macro Avg  & 89& 91 & 81& 82 & 87& 88   & 85& 86\\

\toprule
\end{tabular}
\end{table}

The annotated data set is aggregated from the two annotators on the text level, where each text coming from an FMEA cell or extracted from a presentation slide represents one data instance. 
The annotations of the two annotators are added to this instance. 
Where there is agreement between the annotators, the duplicated annotations are removed. 
When there is a disagreement between the annotators, the annotations are aggregated. 
For the single-stage sequence tagging (SST) method, the different annotations are added to the tokens. 
In the case of the multi-stage sequence tagging (MST) method, the different relations with the same trigger are added as different relations.

The annotated data set is divided into two parts: a testing set and a training set. 
The training data set is further divided into training and validation sets using a five-fold cross-validation approach.
The training set is used to train the model, while the validation set is used for early stopping.
Both the single-stage sequence tagging (SST) and multi-stage sequence tagging (MST) methods are trained using the same training and validation folds and tested using the same testing set.
The performance of the trained models using different language models is summarized in Table~\ref{tab:SST_vs_MST}.
The table provides a comprehensive evaluation of the performance of each method, including the $F1$ score of the multi-label classifier for the SST method. 
Additionally, the table shows the performance of the MST binary classifier for trigger detection, presented in the Trigger $F1$ score.
Also, Trigger grouping model and the multi-label classifier for detecting causes and effects in the case of the MST method.

\begin{table}[h]
\caption{\textbf{Comparison of sequence tagging results using different language models on various test data sets.}
The data sets in include: 
(i) FMEA data set representing texts extracted exclusively from FMEA documents, 
(ii) Slides data set representing texts extracted exclusively from digital presentation slides documents, 
The provided table showcases the mean $F1$ scores achieved by models trained across five folds, employing early stopping based on the development data set.
All the models have shown higher performance on the tagging of the trigger tokens.
In domain pre-training is shown to be effective in increasing the model performance.
}
\label{tab:SST_vs_MST}
\begin{tabular}{l l c c c c c c c c}
\toprule

\multirow{4}{*}{Model} & \multirow{4}{*}{Annotation Type} & \multicolumn{4}{c}{Test Data set} & \multicolumn{2}{c}{Micro Avg} & \multicolumn{2}{c}{Macro Avg} \\
&   &\multicolumn{2}{c}{FMEA}  &\multicolumn{2}{c}{Slides} & \multicolumn{2}{c}{FMEA \&  Slides} & \multicolumn{2}{c}{FMEA \&  Slides} \\
&   & \multicolumn{2}{c}{$F1\%$} & \multicolumn{2}{c}{$F1\% $} & \multicolumn{2}{c}{$F1\% $}&\multicolumn{2}{c}{$F1\% $}\\
&  &  SST  & MST  & SST  & MST   & SST  & MST  & SST  & MST  \\

\midrule

\multirow{5}{*}{BERT}

& Trigger   & 98$\pm$1 & 98$\pm$2   & 75$\pm$4& 81$\pm$2  & 86$\pm$2 &89$\pm$2    & 86$\pm$2&  90$\pm$2  \\
& Cause     & 88$\pm$2 & 88$\pm$4   & 61$\pm$3& 62$\pm$6  & 74$\pm$2 &73$\pm$5    & 74$\pm$2&  75$\pm$5 \\
& Effect    & 74$\pm$9 & 85$\pm$6   & 61$\pm$6& 58$\pm$2  & 67$\pm$7 &69$\pm$3    & 68$\pm$8&  72$\pm$4 \\
& Macro Avg & 87$\pm$4 & 90$\pm$4   & 66$\pm$4& 67$\pm$3  & 76$\pm$4 &77$\pm$3    & 76$\pm$4&  79$\pm$4 \\
& Trigger Grouping & - & 96$\pm$3   &   -     & 97$\pm$4  &   -       &96$\pm$4   &   -     &  96$\pm$4\\
\\
\multirow{5}{*}{BERT UM}

& Trigger    &99$\pm$1&98$\pm$1     &81$\pm$4& 81$\pm$3     & 89$\pm$2 &89$\pm$1    &90$\pm$2&  90$\pm$2\\
& Cause      &92$\pm$5&87$\pm$7     &63$\pm$6& 57$\pm$6     & 76$\pm$4 &71$\pm$4    &78$\pm$6&  72$\pm$6\\
& Effect     &90$\pm$4&89$\pm$3     &67$\pm$6& 57$\pm$4     & 76$\pm$4 &70$\pm$2    &78$\pm$5&  73$\pm$4\\
& Macro Avg  &94$\pm$3&91$\pm$4     &70$\pm$5& 65$\pm$4     & 80$\pm$3 &77$\pm$2    &82$\pm$4&  78$\pm$4\\
& Trigger Grouping& - & 98$\pm$3    &   -    & 100$\pm$0    &   -      &99$\pm$1    & -      &  99$\pm$2\\
\\
\multirow{5}{*}{BERT PMI}

& Trigger   &99$\pm$1&99$\pm$1      &79$\pm$3&82$\pm$ 4     &89$\pm$2 &90$\pm$2     &89$\pm$2&  90$\pm$2\\
& Cause     &89$\pm$5&90$\pm$3      &65$\pm$3&64$\pm$ 6     &76$\pm$3 &75$\pm$4     &77$\pm$4&  77$\pm$4\\
& Effect    &88$\pm$7&91$\pm$5      &71$\pm$4&59$\pm$ 2     &78$\pm$5 &73$\pm$2     &80$\pm$6&  75$\pm$4\\
& Macro Avg &92$\pm$4&93$\pm$3      &72$\pm$3&68$\pm$ 4     &81$\pm$3 &79$\pm$3     &82$\pm$4&  81$\pm$3\\
& Trigger Grouping& -&98$\pm$1      &   -   &100$\pm$1      &   -     &99$\pm$0     &   -    &  99$\pm$1\\
\\
\multirow{5}{*}{MatBERT}

& Trigger   &97$\pm$2&98$\pm$1      &76$\pm$3 &80$\pm$4     &86$\pm$2&89$\pm$2      &86$\pm$2 & 89$\pm$2\\
& Cause     &87$\pm$3&89$\pm$2      &64$\pm$17&66$\pm$7     &75$\pm$8&75$\pm$3      &76$\pm$10& 78$\pm$5\\
& Effect    &82$\pm$8&93$\pm$4      &60$\pm$10&64$\pm$5     &71$\pm$9&79$\pm$2      &71$\pm$9 & 78$\pm$4\\
& Macro Avg &89$\pm$4&93$\pm$2      &67$\pm$10&70$\pm$5     &77$\pm$6&81$\pm$2      &78$\pm$7 & 82$\pm$4\\
& Trigger Grouping& -&95$\pm$4      &   -     &98$\pm$3     &   -    &96$\pm$2      &   -     & 96$\pm$4\\
\\
\multirow{5}{*}{MatBERT UM}

& Trigger   &97$\pm$2&97$\pm$1      &78$\pm$4 &81$\pm$4     &87$\pm$3&88$\pm$2      &88$\pm$3& 89$\pm$2\\
& Cause     &88$\pm$4&86$\pm$8      &66$\pm$11&69$\pm$6     &76$\pm$5&76$\pm$3      &77$\pm$8& 77$\pm$7\\
& Effect    &84$\pm$9&89$\pm$4      &69$\pm$6 &63$\pm$8     &77$\pm$6&75$\pm$3      &76$\pm$6& 76$\pm$6\\
& Macro Avg &90$\pm$4&91$\pm$4      &71$\pm$7 &71$\pm$6     &80$\pm$5&80$\pm$3      &80$\pm$6& 81$\pm$5\\
& Trigger Grouping& -&96$\pm$2      &   -     &98$\pm$2     &   -    &97$\pm$2      &   -    & 97$\pm$2\\
\\
\multirow{5}{*}{MatBERT PMI}

& Trigger   & 96$\pm$1 &98$\pm$1    &76$\pm$5 &82$\pm$4     &86$\pm$3 &90$\pm$2     &86$\pm$3 & 90$\pm$2\\
& Cause     & 83$\pm$8 &88$\pm$4    &67$\pm$7 &65$\pm$4     &74$\pm$1 &75$\pm$3     &75$\pm$8 & 76$\pm$4\\
& Effect    & 86$\pm$10& 94$\pm$3    &65$\pm$10&72$\pm$6     &76$\pm$10&82$\pm$4     &76$\pm$10& 83$\pm$4\\
& Macro Avg & 88$\pm$6 & \textbf{93}$\pm$3    &69$\pm$7 &\textbf{73}$\pm$5     &79$\pm$5 &82$\pm$3     &79$\pm$7 & \textbf{83}$\pm$3\\
& Trigger Grouping& -&93$\pm$4      &   -     & 98$\pm$2    &    -    & 96$\pm$2    &      -  & 96$\pm$3\\
\toprule
\end{tabular}
\end{table}

The results show that both the SST and MST methods perform better in detecting tokens with the label Trigger compared to detecting the tokens with the labels Cause and Effect. 
Additionally, both methods perform better on texts extracted from FMEA documents compared to texts extracted from presentation slides.
Furthermore, both models show an increase in performance when changing the initial language model from BERT to MatBERT without fine-tuning.
The MST method outperforms the SST method in detecting the different labels when using models without in-domain fine-tuning.

The MST method experiences a positive performance impact when using BERT as a language model, texts from FMEA as the testing set, and using UM as a masking objective. 
This performance increases further when using PMI as a masking objective. 
At the same time, the MST method experiences a positive performance impact when using MatBERT as a language model, texts from presentation slides as the testing set, and using UM as a masking objective. 
This performance also increases further when using PMI as a masking objective.

Similarly, the SST method experiences a positive performance impact when using BERT as a language model, texts from presentation slides as the testing set, and using UM as a masking objective. This performance increases further when using PMI as a masking objective.
Surprisingly, the SST method also experiences a positive performance impact when using MatBERT as a language model, texts from FMEA documents as the testing set, and using UM as a masking objective. 
This performance also increases further when using PMI as a masking objective.

\section{Discussion}\label{sec6}

The suggested annotation guidelines have demonstrated their efficiency, as reflected by the high level of agreement between annotators on the annotated dataset.
The results show a higher inter-annotator agreement on the triggers compared to the causes and effects. 
This difference in agreement may be due to the fact that triggers usually consist of only a few tokens, while causes and effects tend to be longer and have less clear-cut and unambiguous boundaries. 
Additionally, the IAA for FMEA is higher than presentation slides. 
This difference in IAA can be attributed to the fact that the content in FMEA documents is typically more structured and organized than in presentation slides. 
FMEA documents often follow a standard format and contain specific sections related to failure modes, effects, and causes. As a result, annotators may find it easier to identify and annotate these sections consistently. 
In contrast, presentation slides often contain unstructured and varied content, with information presented in different formats and in no particular order. 
This variability can make it more challenging for annotators to identify and annotate the relevant content consistently.

Two types of causal information extraction methods, based on single-stage sequence tagging and multi-stage sequence tagging, are also presented in the study. 
The single-stage sequence tagging approach leverages BERT-based language models and supports the extraction of overlapping entities. 
This method is simple to implement as it only requires a multi-label token classifier. 
However, this approach is limited when it comes to extracting enchained and nested relations and disrupted entities. 
The multi-stage sequence tagging addresses these limitations and achieves good performance, sufficient for practical applications with  $F1\_score>90\%$ on FMEA documents and $F1\_score>70\%$ on presentation slides. 
Both methods perform better on texts from FMEA documents than on texts from presentation slides. 
This could be attributed to the same finding from the inter-annotator agreement, as presentation slides are more difficult to annotate. 
Additionally, this could be attributed to the small data set size, as the training data set for presentation slides is less representative of its complexity compared to the FMEA documents.

Choosing a language model that is more aligned with the domain has shown to improve the performance of both the SST and MST methods. However, the effect of in-domain fine-tuning on the model performance is dependent on the test data set (i.e., FMEA or presentation slides) and the initial language model.
In-domain fine-tuning can be an effective method for further improving the performance of the SST and MST methods.
The performance improvement seems to differ for the different entity types, with the F1 score for effect increasing more than the F1 score for cause.
This could indicate that effect entities are more domain specific than the cause entities.
However, the effectiveness of in-domain fine-tuning is highly dependent on the initial language model and the data set used for fine-tuning.
Overall, our findings suggest that careful consideration of the language model and fine-tuning approach is important for achieving high-quality causal information extraction from unstructured text data in the semiconductor industry.

\subsection*{Limitations and opportunities}

In this paper, we are able to demonstrate the effectiveness of a multi-stage sequence tagging (MST) approach in industrial settings. 
In the MST approach, the trigger is detected first and the cause and effect are then identified on a given trigger.
However, the method uses an OCR for extracting texts from presentation slides. 
Therefore, it is limited to the extraction of textualized causal relations in presentation slides and cannot process causal information expressed through visual elements like arrows, color encoding, etc. 
This limitation is of high relevance as presentation slides are created to be viewed and not be simply read, since a great amount of their information is in visualized form. 
This information is lost when converting the content into plain texts. 
To address this limitation, it is necessary to explore hybrid approaches in which NLP approaches are augmented with computer vision methods.
Furthermore, the content extraction via OCR is prone to errors.
Especially tabular structures, line breaks, hyphenated words, small fonts and images with blurry texts can pose problems to the OCR.

Besides, the multi-stage approach only allows to extract causal relations which are expressed through explicit linguistic triggers.
Causes and effects without an explicit trigger cannot be detected, since the trigger is given as an input to detect the cause and effect in the second stage of the NER. 
Even though the chances for two elements to be perceived as causally-related by the annotators is lower the further away they are from each other in the text ~\citep{riaz2010another}, it is possible to have causal relations spanning over a few sentences or even a whole document. 
To extract such relations inter-sentential or intra-document relation extraction is necessary.
Our method focuses on intra-text causality.
One text corresponds to the content of one text box or bullet point in terms of presentation slides or one cell in terms of content from tabular formats.
This means that this method is not able to detect causal relations between elements from different text boxes or cells.
Another drawback of this approach is its dependence on human annotations. This reliance raises not just considerations of resource allocation - given the time and cost associated with large-scale human annotation - but also presents the daunting task of resolving varying interpretations of causality among different annotators.
While the suggested guidelines hold the potential to serve as prompts for sophisticated language models in the future, their practical implementation is yet to be fully assessed. 
It is essential to undertake more in-depth research to explore the viability of this approach, particularly with a keen focus on addressing potential information security concerns.

\section{Conclusion}\label{Conclusion}

This paper provides valuable insights into the development of automated methods for causal information extraction from semi structured and unstructured documents in the semiconductor manufacturing industry.
The proposed method is expected to bridge the gap between industry and academia, increasing the availability of causal domain knowledge for downstream tasks in the industry. 
The effectiveness of the suggested annotation guidelines is evident from the considerable level of agreement between annotators in relation to the annotated dataset.
Also, the study demonstrates the effectiveness of two types of causal information extraction methods, single-stage sequence tagging and multi-stage sequence tagging, using actual industrial documents. 
The adapted multi-stage sequence tagging based method can effectively capture complex entities and relations, including enchained and nested relations and disrupted entities.
This makes it a more optimal choice for extracting causal information.
Moreover, the paper provides guidance for practitioners working in different industries with similar types of documents.
Namely, the proposed annotation guidelines can be considered as a solid basis for annotating causal information for other domains and other types of documents.  
In addition, the study highlights the importance of representation learning on downstream tasks. Choosing a language model that is more aligned with the domain, and in-domain fine-tuning has been shown to significantly improve the performance of the SST and MST methods.
Overall, this study contributes to the body of research that aims to develop more accurate and efficient automated methods for causal information extraction from unstructured text data in the semiconductor industry, and provides valuable insights that can be applied to other industries with similar types of documents.

\section*{Data availability statement}
This research draws upon valuable industry data from a semiconductor manufacturing company. However, due to confidentiality agreements, the specific data cannot be publicly shared.

\section*{Competing interest statement}

The authors have no competing interests to declare that are relevant to the content of this article.

\section*{Acknowledgment statement}

Part of this work is  conducted under the framework of the EdgeAI “Edge AI Technologies for Optimised Performance Embedded Processing” project has received funding from Chips Joint Undertaking (Chips JU), under grant agreement No 101097300. The Chips JU receives support from the European Union’s Horizon Europe research and innovation program and Austria, Belgium, France, Greece, Italy, Latvia, Netherlands, Norway.

\bibliography{sn-bibliography}

\end{document}